\documentclass[final]{cvpr}

\usepackage{times}
\usepackage{epsfig}
\usepackage{graphicx}
\usepackage{amsmath}
\usepackage{amssymb}
\usepackage{color}
\usepackage{multirow, booktabs}
\usepackage[normalem]{ulem}
\usepackage{xcolor}
\usepackage{nopageno}
\usepackage{balance}
\newcommand\bluesout{\bgroup\markoverwith{\textcolor{blue}{\rule[0.5ex]{2pt}{0.4pt}}}\ULon}
\newcommand\cyansout{\bgroup\markoverwith{\textcolor{cyan}{\rule[0.5ex]{2pt}{0.4pt}}}\ULon}
\usepackage{subfigure}

\usepackage[pagebackref=true,breaklinks=true,colorlinks,bookmarks=false]{hyperref}

\def\cvprPaperID{8215} 
\def\confYear{CVPR 2021}

\begin{document}

\title{CorrNet3D: Unsupervised End-to-end Learning of Dense Correspondence \\for 3D Point Clouds}

\author{ 
\and

Yiming Zeng$^{1*}$\quad Yue Qian$^1$\quad Zhiyu Zhu$^1$\quad Junhui Hou$^{1\dagger}$\quad Hui Yuan$^2$ \quad Ying He$^3$\\
$^1$
City University of Hong Kong~~ 
$^2$
Shandong University~~ 
$^3$
Nanyang Technological University\\
{\tt\small \{ym.zeng, yueqian4-c, zhiyuzhu2-c\}@my.cityu.edu.hk, jh.hou@cityu.edu.hk,  yhe@ntu.edu.sg}
}

\maketitle
\footnotetext[1]{This work was supported by the HK RGC Grant CityU 11202320, the NSFC Grant 61871342, and Singapore MOE Grant 20/20.}
\footnotetext[2]{Corresponding author.}

\begin{abstract}
Motivated by the intuition
that one can transform two aligned point clouds to each other more easily and meaningfully than a misaligned
pair, we propose CorrNet3D -- the first unsupervised and end-to-end deep learning-based framework -- to drive the learning of dense correspondence between 3D shapes by means of deformation-like reconstruction to overcome the need for annotated data.
Specifically, CorrNet3D consists of a deep feature embedding module and two novel modules called correspondence indicator and symmetric deformer. Feeding a pair of raw point clouds, our model first learns the pointwise features and passes them into the indicator to generate a learnable correspondence matrix used to permute the input pair. The symmetric deformer, with an additional regularized loss, transforms the two permuted point clouds to each other to drive the unsupervised learning of the correspondence. The extensive experiments on both synthetic and real-world datasets of rigid and non-rigid 3D shapes show our CorrNet3D outperforms state-of-the-art methods to a large extent, including those taking meshes as input. 
CorrNet3D is a flexible framework in that it can be easily adapted to supervised learning if annotated data are available. The source code and pre-trained model will be available at \href{https://github.com/ZENGYIMING-EAMON/CorrNet3D.git}{https://github.com/ZENGYIMING-EAMON/CorrNet3D.git}.
\end{abstract}

\section{Introduction}

Owing to the flexibility and efficiency in representing 3D objects/scenes as well as the recent advances in 3D sensing technology, 3D point clouds have been widely adopted in various applications,
e.g., immersive communication~\cite{apostolopoulos2012road}, autonomous driving~\cite{raviteja2020introduction}, AR/VR~\cite{silva2003introduction}, etc. 
Since each camera/scanner produces a point cloud in its own camera space 
rather than the object space, there is no correspondence between two point clouds (even they represent the same object), 
which poses great challenges for downstream processing and analysis, such as 
motion transfer~\cite{sun2020human},  shape editing~\cite{liu2019deep}, dynamic point cloud compression~\cite{huang20193d}, object recognition~\cite{arnold2019survey}, shape retrieval~\cite{gezawa2020review}, 
surface reconstruction \cite{bronstein2017geometric}, and many others.

Building dense shape correspondence is a fundamental and challenging problem in computer vision and digital geometry processing. There are a considerable number of methods proposed, which   
can be roughly classified into two categories: model-based~\cite{bronstein2006generalized,huang2008non,tevs2011intrinsic,ovsjanikov2012functional} and data-driven~\cite{litany2017deep,donati2020deep,corman2014supervised}.
The model-based methods usually use handcrafted features to optimize pre-defined processes. The recent deep learning-based methods train their neural networks in a data-driven manner and improve the performance to a large extent. However, the existing methods either require a large amount of annotated data which are difficult to obtain or assume the connectivity information is available in the input data, i.e., polygonal meshes. This paper focuses on unsupervised learning of dense correspondence between non-rigid 3D shapes in the form of 3D point clouds, but the proposed method can also be used for rigid 3D shapes. 

\begin{figure}[t]
  \centering
  \includegraphics[width=0.5\textwidth]{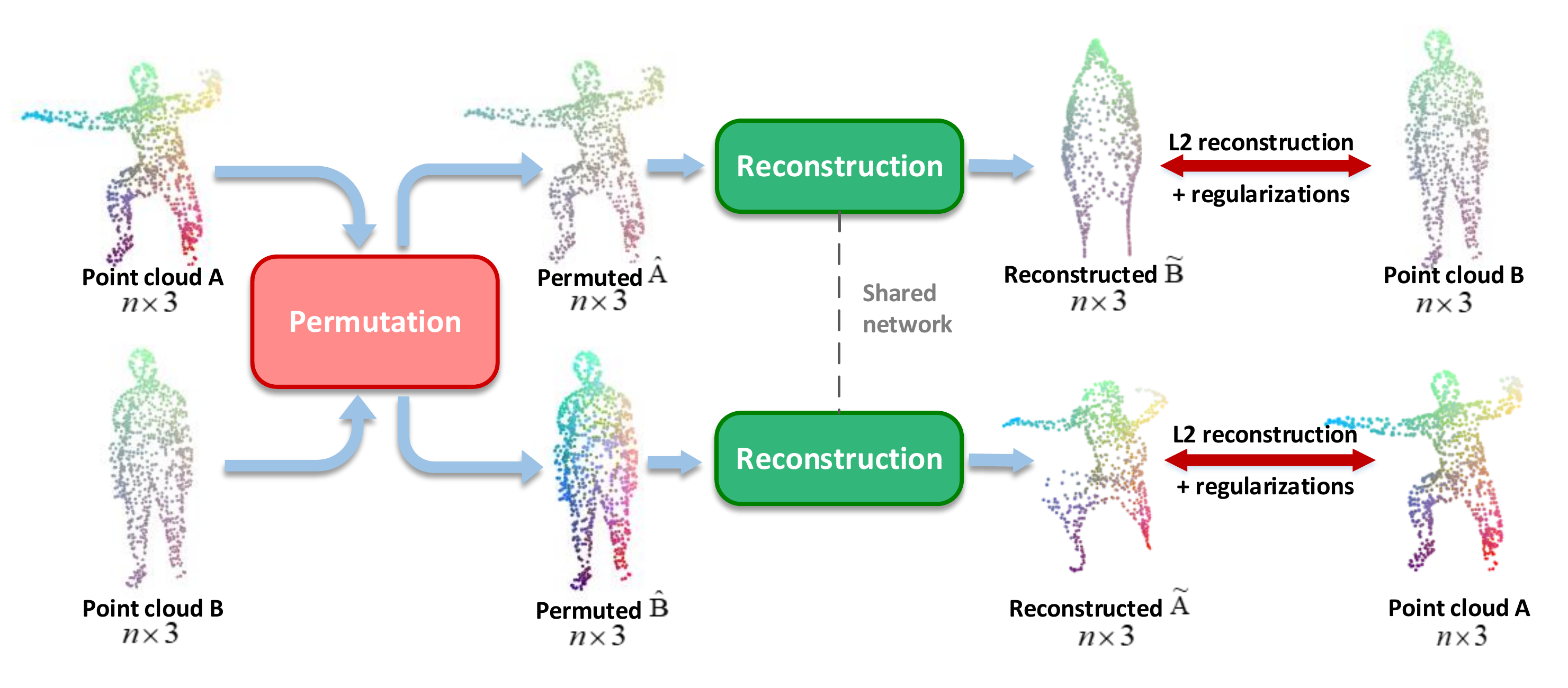}
  \caption{Illustration of the motivation of our unsupervised deep learning-based framework for computing dense correspondence between two point clouds. 
}
  \label{summary}
\end{figure}

\textbf{Motivation}.   
Let $\mathbf{A}\in\mathbb{R}^{n\times 3}$ and $\mathbf{B}\in\mathbb{R}^{n\times 3}$ be the two point clouds to be corresponded\footnote{Note that the points are randomly stacked to form a matrix.}, where 
$\mathbf{a}_i=\{x_i,y_i,z_i\}$, $1\leq i\leq n$ and $\mathbf{b}_j=\{x'_j,y'_j,z'_j\}$, $1\leq j\leq n$ are the $i$-th and $j$-th 3D points of $\mathbf{A}$ and $\mathbf{B}$, respectively. 
Fig.~\ref{summary} illustrates our motivation: 
if $\mathbf{A}$ and $\mathbf{B}$ are well aligned, it is easier to transform one model to the other. 
More precisely, denote 
by $\widehat{\mathbf{A}}\in\mathbb{R}^{n\times 3}$ and $\widehat{\mathbf{B}}\in\mathbb{R}^{n\times 3}$ the re-ordered $\mathbf{A}$ and $\mathbf{B}$ via a designed permutation process, respectively.  
With a designed reconstruction process, it is expected that we can reconstruct $\mathbf{A}$ (resp. $\mathbf{B}$) from $\widehat{\mathbf{B}}$ (resp. $\widehat{\mathbf{A}}$) more easily and meaningfully than the manner of reconstructing $\mathbf{A}$ (resp. $\mathbf{B}$) from $\mathbf{B}$ (resp. $\mathbf{A}$) directly. 
Therefore, we can minimize the reconstruction error from $\widehat{\mathbf{B}}$ (resp. $\widehat{\mathbf{A}}$) to $\mathbf{A}$  (resp. $\mathbf{B}$) to drive the learning of the permutation process, which implicitly encodes the dense correspondence between $\mathbf{A}$ and $\mathbf{B}$. 

Based on the above intuitive understanding, we propose the \textit{first unsupervised and end-to-end} deep learning-based framework for point clouds.
Technically, we propose a novel correspondence indicator and a deformation-like reconstruction module to achieve the permutation and reconstruction processes, respectively. To be specific, the correspondence indicator, fed with point-wise high-dimensional feature representations of the input point clouds learned by a hierarchical feature embedding module,   
generates a permutation matrix, which explicitly encodes the point-to-point correspondence. 
During training, the deformation-like reconstruction module receives the aligned point clouds and the global semantic features of inputs to reconstruct each  other by optimizing the reconstruction error and additional regularization terms to drive the learning of the permutation matrix.

In summary, we make the following contributions.
\vspace{-0.5em}
\begin{enumerate}
    \item We propose the first unsupervised deep learning framework for building dense correspondence between point clouds in an end-to-end manner.
    \vspace{-0.25cm}
    \item We propose two novel modules, i.e., the correspondence indicator with the efficient DeSmooth module, the symmetric deformation module, as well as a novel loss function. 
    \vspace{-0.25cm}
    \item We show that CorrNet3D can be adapted to both unsupervised and supervised conditions, and handle both non-rigid and rigid shapes well. 
    \vspace{-0.5em}
    \item 
    We experimentally demonstrate the significant superiority of CorrNet3D over state-of-the-art methods. Especially, 
    CorrNet3D even outperforms the method taking 3D meshes as input. 
\end{enumerate}  
\section{Related Work}



\subsection{Deep Learning for Point Clouds}
 Unlike well-developed deep convolution neural network (CNN) techniques for 2D images/videos, deep learning based point cloud processing is more challenging and still in the infant stage, due to its irregular and unorder characteristics. 
PointNet~\cite{qi2017pointnet} and PointNet++~\cite{qi2017pointnet++} are the pioneering works 
and verify the effectiveness of multi-layer perceptrons (MLPs) in learning point cloud features.
DGCNN~\cite{dgcnn} uses a dynamic graph to aggregate neighborhood information in each layer, and the selection of neighbours is based on feature distances.
DCG \cite{wang2019DCG} further boosts  DGCNN by encoding additional local connections in coarse-to-fine manner.
Volumetric-based methods~\cite{ModelNet, VoxNet,OctNet, OCNN,KD-Net} apply 3D CNNs to process voxelized point clouds; however, they suffer from high computational costs and inevitable quantization errors.
In the meantime, inspired by the FoldingNet~\cite{yang2018foldingnet}, which learns to deform pre-defined 2D regular grids into 3D shapes,
some deformation-based frameworks, such as AtlasNet \cite{groueix2018papier} and 3D-Coded \cite{groueix20183d}, were proposed, 
which deform 
a fixed templates (e.g., 2D grid or 3D human mesh) to reconstruct the input point cloud or mesh. Please refer to \cite{guo2020survey} for the comprehensive survey on deep learning-based point cloud processing.




\begin{figure*}[t]
  \centering
  \includegraphics[width=0.95\textwidth]{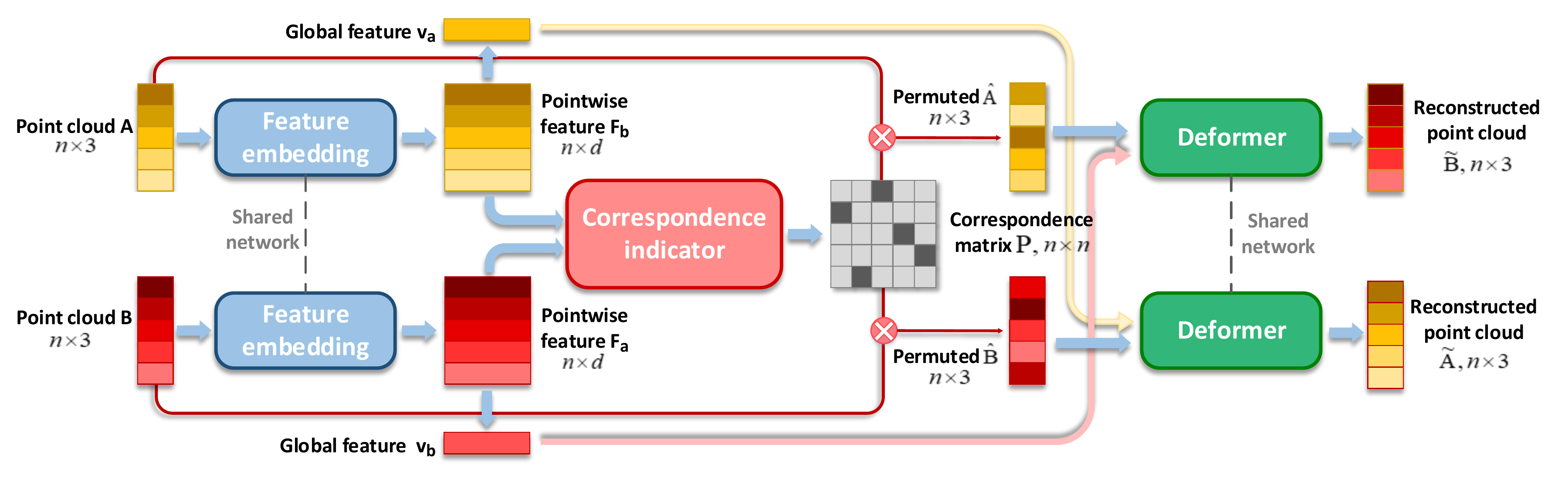}
  \caption{The flowchart of CorrNet3D, an \textit{unsupervised} and \textit{end-to-end} deep learning framework, which aims to obtain a matrix $\mathbf{P}$, which explicitly indicates the correspondence between any two points. We first 
  represent 
  $\mathbf{A}$ and $\mathbf{B}$ with high-dimensional point-wise features $\mathbf{F}_a$ and $\mathbf{F}_b$ as well as the global features $\mathbf{v}_a$ and $\mathbf{v}_b$. 
Then the correspondence indicator with a novel DeSmooth module takes $\mathbf{F}_a$ and $\mathbf{F}_b$ as input to regress $\mathbf{P}$.
To drive the unsupervised learning of $\mathbf{P}$, two symmetric deformers with shared parameters 
takes the $\mathbf{A}$, $\mathbf{B}$, $\mathbf{P}$ and $\mathbf{v}_a$,  and $\mathbf{v}_b$ as inputs to generate the reconstructed point clouds $\mathbf{\widetilde{A}}$ and $\mathbf{\widetilde{B}}$ in the deformation-like manner. CorrNet3D is trained with the reconstruction loss and additional regularization terms on $\mathbf{P}$.}
  \label{models}
\end{figure*}

\subsection{Non-rigid Shape Correspondence}\label{nonrigidsm}
Non-rigid shape correspondence or matching aims to find the point-to-point correspondence of two deformable 3D shapes. 
As an active research area in computer vision and graphics, many methods have been proposed, and one may refer to the  surveys~\cite{van2011survey,sahilliouglu2020recent,biasotti2016recent} for a comprehensive understanding. Here we briefly introduce one stream, i.e., functional map (FM)-based methods
~\cite{ovsjanikov2012functional} which are compared in this paper. 
Specifically, this kind of methods first performs spectral analysis on 3D meshes to construct an FM, and then optimizes a least-squares problem to convert the resulting FM to 
point-to-point correspondence under the assumption of non-rigid 
but isometric deformation.
To overcome the deficiency of solving the optimization problem, Litany \textit{et al.}~\cite{litany2017deep} proposed 
deep FM, which integrates the module of computing FM into 
a deep neural network. 
However, both original FM-based methods and deep FM use the handcrafted SHOT descriptors~\cite{salti2014shot}
, which may limit their performance. 
Based on deep FM, Halimi et al. \cite{halimi2019unsupervised} proposed to minimize the surface distortion between meshes to drive the correspondence learning process. 
Instead of using handcrafted descriptors, 
the most recent deep learning-based method named deep geometric functional map (DeepGFM) \cite{donati2020deep} 
employs KPConv\cite{thomas2019kpconv} to achieve a data-driven feature representation. 
Although DeepGFM can achieve state-of-the-art performance, it is only applicable for 3D meshes, and its training has to be supervised by ground-truth FMs, whose construction requires ground-truth correspondence. Moreover, additional post-processing is necessary to obtain the final correspondence. 
 Recently, FlowNet3D \cite{liu2019flownet3d} was designed to 
directly learn scene flows from two 
consecutive point clouds in the form of depth images. 
To some extent, it can also be used for indicating correspondence, i.e., adding the estimated flow to one point cloud, and then seeking the closest point in the other one. 
However, such a simple extension may result in serious many-to-one correspondence. 
Moreover, due to the specific application scenario, FlowNet3D only utilizes the neighborhood information based on the Euclidean distance in two frames,
making it not applicable to 3D shapes with serious deformation. 
Groueix \textit{et al}. \cite{groueix2019unsupervised} proposed a self-supervised approach to achieve deep surface deformation for shapes in the same category, in which the semantic labels from a small set of segmented shapes are transferred to unlabeled data. This work has potential on shape matching.
In our experiment, we slightly modified the loss function of FlowNet3D to produce an improved unsupervised model for correspondence prediction, which is adopted as a baseline method for comparisons.



\subsection{Rigid Shape Matching}

Rigid shape matching or registration aims to obtain a rotation matrix $\mathbf{R}\in\mathbb{R}^{3\times 3}$ 
and a translation vector $\mathbf{t}\in\mathbb{R}^{3\times 1}$ to align two rigid 3D shapes. Over the past decades, a considerable number of methods have been proposed. Please refer to \cite{Tam2013registrationsurey} for the comprehensive survey of traditional methods for rigid 3D shape registration.
Recently, some deep learning-based methods 
were proposed. 
For example, PointNetLK \cite{aoki2019pointnetlk} 
utilizes PointNet \cite{qi2017pointnet} to extract global features for two point clouds separately and then estimate $\mathbf{R}$ and $\mathbf{t}$. 
DCP \cite{wang2019deep} introduces a transformer \cite{NIPS2017_7181} to solve the seq-2-seq problem, where the point-wise correspondence and ($\mathbf{R}$, $\mathbf{t}$) are simultaneously estimated.
The recent work RPMNet \cite{yew2020rpm} 
adopts the Sinkhorn layer \cite{mena2018learning} to 
get the correspondence information and  weighted 
SVD \cite{golub1971singular} to compute $\mathbf{R}$ and $\mathbf{t}$.
Note that all these learning-based methods require ground-truth rotations and translations as supervision or even additional post-processing.
\section{Proposed Framework}

\subsection{Overview}

As illustrated in Fig.~\ref{models}, 
our CorrNet3D mainly consists of three modules: feature embedding, correspondence indicator, and  symmetric deformer. Specifically,
we first pass paired input point clouds into the shared feature embedding module to generate point-wise high-dimensional feature embeddings
$\mathbf{F}_a\in\mathbb{R}^{n\times d}$ and $\mathbf{F}_b\in\mathbb{R}^{n\times d}$ with $d$ being the feature dimension,  which encode their local geometric structures, respectively,  
and global feature vectors $\mathbf{v}_a\in\mathbb{R}^d$ and $\mathbf{v}_b\in\mathbb{R}^d$, which encode their shape information, respectively. 
Then we predict a 
correspondence matrix $\mathbf{P}\in\mathbb{R}^{n\times n}$ by feeding $\mathbf{F}_a$ and $\mathbf{F}_b$ into 
the correspondence indicator, where the the $(i,j)$-th element $p_{ij}=1$ indicates the point $\mathbf{a}_i$ corresponds to $\mathbf{b}_j$. 
To drive the learning of $\mathbf{P}$ in an unsupervised manner, we 
propose the symmetric deformer  
in which 
we utilize $\mathbf{v}_b$ (resp. $\mathbf{v}_a$ ) and the permuted point cloud $\widehat{\mathbf{A}}$ (resp. $\widehat{\mathbf{B}}$) to reconstruct $\mathbf{B}$ (resp. $\mathbf{A}$).
CorrNet3D is end-to-end trained 
by directly minimizing $\|\mathbf{A}-\widetilde{\mathbf{A}}\|_F^2+\|\mathbf{B}-\widetilde{\mathbf{B}}\|_F^2+\lambda\mathcal{R}(\mathbf{P})$, 
where $\widetilde{\mathbf{A}}\in\mathbb{R}^{n\times 3}$ and $\widetilde{\mathbf{B}}\in\mathbb{R}^{n\times 3}$ are the reconstructed point clouds, $\|\cdot\|_F$ is the Frobenious norm of a matrix, $\lambda>0$ is the penalty parameter, and $\mathcal{R}(\mathbf{P})$ stands for the regularization on $\mathbf{P}$. 

\textbf{Remark}. The proposed CorrNet3D is fundamentally different from the existing works \cite{yew2020rpm},\cite{donati2020deep}, as the correspondence matrix is driven from the perspective of deformation-like reconstruction, 
rather than the ground-truth correspondence or the well-known functional maps. In addition, CorrNet3D is able to work as a supervised model by removing the deformation module and employing ground-truth correspondence to supervise the learning of $\mathbf{P}$.  
In the experiment section, we demonstrate the significant advantage of CorrNet3D under both unsupervised and supervised scenarios. 

\subsection{Feature Embedding} 
We use a shared DNN-based feature learner, namely DGCNN \cite{dgcnn},  to embed $\mathbf{A}$  and $\mathbf{B}$ to a high-dimensional feature space 
in a hierarchical manner. Note that other advanced feature representation methods \cite{sun2020pointgrow}, \cite{boulch2020convpoint} can also be used to further boost performance.
To be specific, DGCNN consists of several layers named EdgeConv.  
For the $i$-th point, we first calculate the Euclidean distance between features 
to determine the set of its $k$ nearest neighbours 
denoted by $\Omega_i^l$. 
Then, we apply  
an MLP~\cite{hastie2009elements} 
followed by a max-pooling operator $\square$ to obtain a new feature representation  $\mathbf{f}_i^{l+1}=\square_{\mathbf{f}_j^l\in \Omega_i^l}\mathcal{M}_l(\mathbf{f}_i^l, \mathbf{f}_j^l-\mathbf{f}_i^l)$, where $\mathbf{f}_i^{l}\in\mathbb{R}^{1\times d}$ be the feature representation of point $i$ fed into the $l$-th EdgeConv.  $\mathbf{f}_i^{l+1}$ is capable of capturing the local geometry structure of point $i$.  
After $L$ EdgeConv layers, we can obtain the final point-wise features $\mathbf{F}_a\in\mathbb{R}^{n\times d}$ for $\mathbf{A}$ and $\mathbf{F}_b\in\mathbb{R}^{n\times d}$ for $\mathbf{B}$. By applying another max-avg-pooling operator, the global feature vectors for A and B could be accordingly obtained denoted as   $\mathbf{v}_a\in\mathbb{R}^d$ and $\mathbf{v}_b\in\mathbb{R}^d$, respectively. 
Please refer to \cite{dgcnn} for more details about DGCNN. 



\subsection{Correspondence Indicator} 

\begin{figure}[t]
  \centering
  \includegraphics[width=0.4\textwidth]{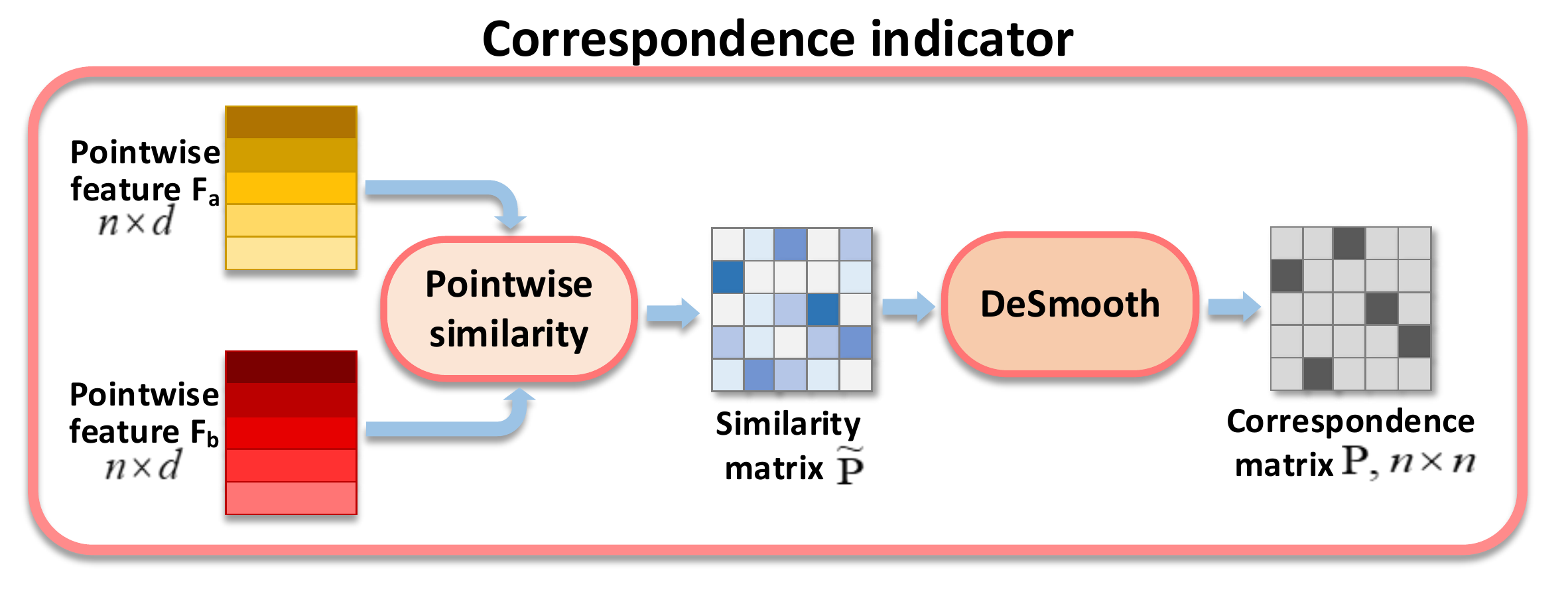}
  \caption{ Illustration of the correspondence indicator, which predicts a correspondence matrix $\mathbf{P}$ by taking  pointwise features $\mathbf{F}_a$ and $\mathbf{F}_b$ as input.}
  \label{fig:ci}
\end{figure}

Our correspondence indicator module aims to learn a correspondence matrix $\mathbf{P}\in\mathbb{R}^{n\times n}$ to explicitly indicate the correspondence between any two points of $\mathbf{A}$ and $\mathbf{B}$. Ideally, $\mathbf{P}$ should be a permutation matrix that is binary and orthogonal, and iff $p_{ij}=1$, point $\mathbf{a}_i$ corresponds to point $\mathbf{b}_j$. 
However, such a permutation matrix is non-differentiable, making it hard to optimize in a deep learning framework. Alternatively, we regress an approximate doubly stochastic matrix instead, which is differentiable and close to binary. Moreover,  there is only a single prominent element dominating each row and column. During inference, we quantize $\mathbf{P}$ to an exact binary matrix $\widehat{\mathbf{P}}$. 

As shown in Fig. \ref{fig:ci}, to learn $\mathbf{P}$, 
we first measure the similarity between points in the high-dimensional feature space 
in the inverse distance sense, i.e.,  
\begin{equation}
    \widetilde{p}_{ij}=\frac{1}{\|\mathbf{f}_{a,i}-\mathbf{f}_{b,j}\|_2},\label{eq1}
\end{equation}
where $\widetilde{p}_{ij}\subset\widetilde{\mathbf{P}}\in\mathbb{R}^{n\times n}$, and $\mathbf{f}_{a,i}$ and $\mathbf{f}_{b,j}\in\mathbb{R}^{1\times d}$ are the $i$-th and $j$-th point-wise features corresponding to $\mathbf{a}_i$ and $\mathbf{b}_j$, respectively. 
However, the resulting $\widetilde{\mathbf{P}}$ is far away from realizing correspondence. 
To further enhance $\widetilde{\mathbf{P}}$, one can simply adopt 
 Sinkhorn layers \cite{mena2018learning}, which perform the softmax operation on in column-wise and row-wise iteratively and alternatively, as done in 
\cite{yew2020rpm}; however, the efficiency of Sinkhorn layers is low due to the iterative manner. 
To tackle this issue, we propose a novel DeSmooth module to improve $\widetilde{\mathbf{P}}$. 

\textbf{DeSmooth Module}.
Assume that 
$\widetilde{p}_{ij}$ 
generally obeys a series of Gaussian distributions\footnote{As $\widetilde{p}_{ij}$ is non-negative, it does not follow a strict Gaussian distribution. However, we found such an assumption still works well in practice}: 
$\widetilde{p}_{ij}\sim \mathcal{N}\left(\mu_i ,\sigma_i ^2 \right)$,  where $\mu_i$ and $\sigma_i$ are the mean and standard deviation of the $i$-th row of $\widetilde{\mathbf{P}}$.
We first normalize $\widetilde{p}_{ij}$ in row-wise, i.e., $z_{ij}=\frac{\widetilde{p}_{ij}-\mu_i}{\sigma_i}$. 
Accordingly, $z_{ij}$ follows a  
standard normal distribution 
$z_{ij}=\frac{\widetilde{p}_{ij}-\mu_i}{\sigma_i}\sim \mathcal{N}\left(0 ,1 \right)$.
Give a prior ratio $t$ to $z_{ij}$, we have
\begin{equation}
  \label{eqn:mss}
  \widetilde{z}_{ij} = t\cdot z_{ij} \sim \mathcal{N}\left( 0, t \right).
\end{equation}
For $i$-th row of $\widetilde{\mathbf{Z}}\in\mathbb{R}^{n\times n}$ ($\widetilde{z}_{ij}\subset\widetilde{\mathbf{Z}}$), we compute the number of elements whose values are 
not less than a threshold $\tau$, i.e.,
\begin{equation}
  \label{eqn:di}
  c_i=\#\left\{ \widetilde{z}_{ij}|\widetilde{z}_{ij}\geqslant \tau,j=1,\cdots ,n \right\} 
\end{equation}
where $\#\{\cdot\}$ denotes the cardinality.
We expect the value of $c_i$ 
to be close to 1, which means that in each row there's a high probability that only a single element dominates the row. 

The $n$ rows of $\widetilde{\mathbf{Z}}$ could be thought of as 
$n$ i.i.d events, and thus the set $\mathbf{c}=\left\{c_i|i=1,\cdots,n\right\}$ also follows a Gaussian distribution 
with the expectation $\mu_c$ and variance $\sigma_c$ 
depending 
on the prior ratio $t$. 
Therefore, According to the three-sigma rule \cite{pukelsheim1994three}, we can set a proper $t$ to control the bound $[\mu_c-3\sigma_c, \mu_c+3\sigma_c]$ to be centered around 1, such that the aforementioned expectation on a feasible correspondence matrix can be realized. 
Finally, we apply the softmax operation on $\widetilde{z}_{ij}$ again and obtain the correspondence matrix $\mathbf{P}$, i.e.,
\begin{equation}
  p_{ij}=\frac{e^{\widetilde{z}_{ij}}}{\sum_{j=1}^n{e^{\widetilde{z}_{ij}}}}. 
\end{equation}
The advantage of our DeSmooth over Sinkhorn layers is also experimentally demonstrated in Sec. \ref{ablation_}. 


\subsection{Symmetric Deformer}\label{Sec_deformer}

\begin{figure}[t]
  \centering
  \includegraphics[width=0.4\textwidth]{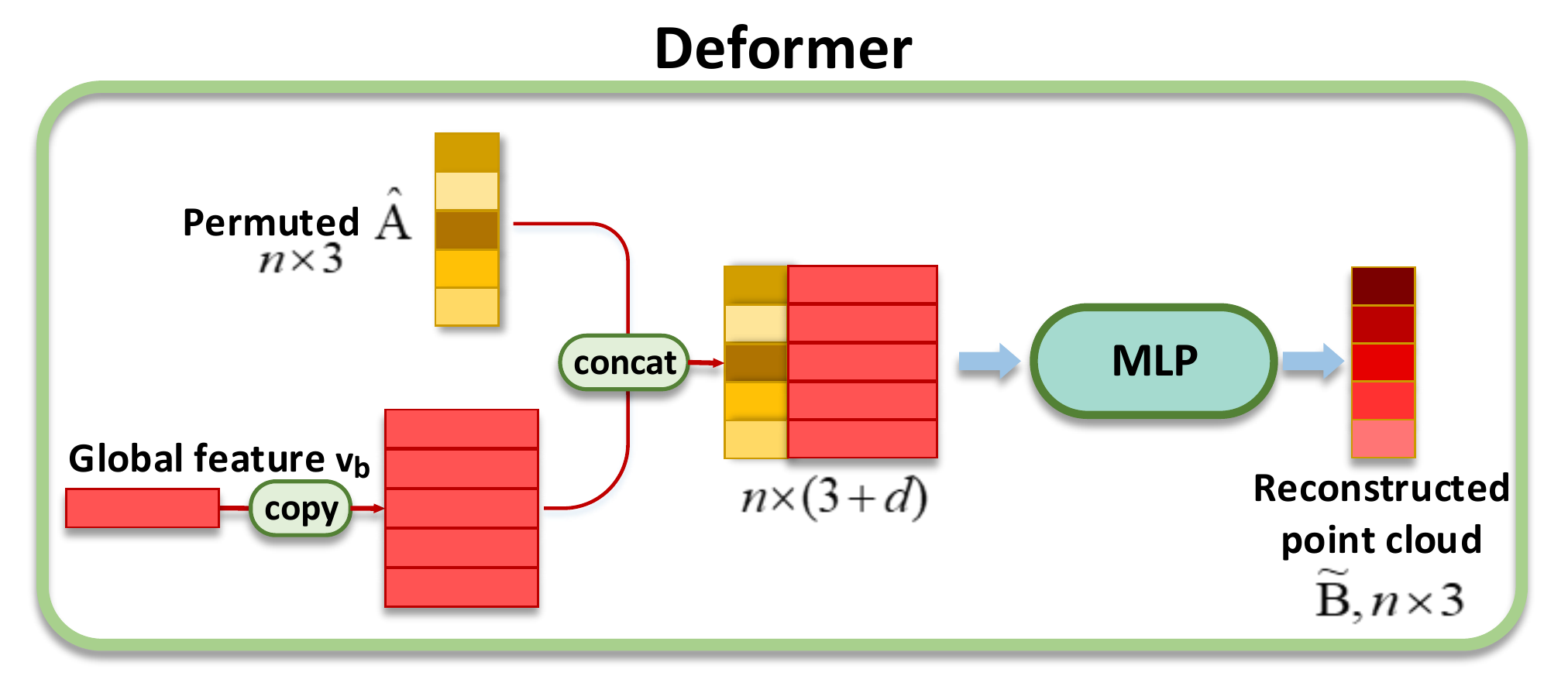}
  \caption{Illustration of one branch of the symmetric deformer. 
  The global feature $\mathbf{v}_b$  encoding the shape information of  $\mathbf{B}$ is concatenated to the 3D coordinate of each point of $\widehat{\mathbf{A}}$, which are fed into an MLP to reconstruct $\mathbf{B}$ from a deformation perspective. 
  }
  \label{ci}
\end{figure}

Given ground-truth correspondence between $\mathbf{A}$ and $\mathbf{B}$, the modules for learning 
$\mathbf{P}$ can be easily trained in a supervised manner. 
Annotating large amount of data is, however, costly and time-consuming. Inspired by recent deformation-based methods, such as FoldingNet~\cite{yang2018foldingnet} and AtlasNet~\cite{groueix2018papier}, which feed a pre-defined 2D grid appended with a global feature vector to a network to reconstruct 3D shapes, 
we propose symmetric deformer shown in Fig. \ref{ci}, which reconstruct $\mathbf{A}$ and $\mathbf{B}$ in a deformation-like fashion to achieve the \textit{unsupervised} learning of $\mathbf{P}$.  
We link the matrix $\mathbf{P}$ with the deformation module based on the aforementioned intuition, i.e.,  it is  more  easily  and  meaningfully to transform two aligned point clouds to each other  than  a  misaligned  pair. 
Specifically,
we first permute the input point clouds using the learned $\mathbf{P}$ approximate to a permutation matrix, i.e.,  
\begin{equation}
    \widehat{\mathbf{A}}=\mathbf{P}^\textsf{T}\mathbf{A},~~~~ 
    \widehat{\mathbf{B}}=\mathbf{P}\mathbf{B}.
    \end{equation}

The resulting $\widehat{\mathbf{A}}$ (resp. $\widehat{\mathbf{B}}$) is approximately aligned with $\mathbf{B}$ (resp. $\mathbf{A}$). Then, we deform $\widehat{\mathbf{A}}$ to $\mathbf{B}$ and $\widehat{\mathbf{B}}$ to $\mathbf{A}$,
by respectively utilizing their overall shape information encoded in  the learned global feature vectors 
$\mathbf{v}_a$ and $\mathbf{v}_b$.
Technically, we concatenate $\mathbf{v}_a$ (resp. $\mathbf{v}_b$) to each point of $\widehat{\mathbf{B}}$ (resp. $\widehat{\mathbf{A}}$) and pass the extended points to a network consisting of MLP layers, leading to reconstructed point clouds $\widetilde{\mathbf{A}}$ (resp. $\widetilde{\mathbf{B}}$).  See Sec.\ref{ablation_} for the experimental validation towards the effectiveness and the deformation behavior of this module. 

\subsection{Unsupervised Loss Function}
To train the proposed CorrNet3D end-to-end,  we promote 
$\widetilde{\mathbf{A}}$ and $\widetilde{\mathbf{B}}$ to be close to $\mathbf{A}$ and $\mathbf{B}$, respectively, which is achieved by   
    \begin{equation}
  \label{recerror}
  \mathcal{L}_{rec}\left(\widetilde{\mathbf{A}},\widetilde{\mathbf{B}} \right) =\left\|\mathbf{A}-\mathbf{\widetilde{A}}\right\|_F^2+\left\|\mathbf{B}-\widetilde{\mathbf{B}}\right\|_F^2.
\end{equation}
Benefiting from the alignment operation involved in CorrNet3D, we are allowed to use such a point-to-point reconstruction loss, which is easier to optimize than the commonly-used CD loss, thus producing better performance. 
See the ablation study.  


In addition to the reconstruction loss, 
we also propose another two terms to regularize the learning of the correspondence matrix $\mathbf{P}$. 
The first regularization term is defined as 
\begin{equation}
       \label{lrank}
 \mathcal{L}_{perm}(\mathbf{P})=\left\|\mathbf{PP}^\textsf{T}-\mathbf{I}_n\right\|_F^2,
\end{equation}
where $\mathbf{I}$ is the identity matrix of size $n\times n$. Such a term  encourages 
$\mathbf{P}$ to be close to a permutation matrix to eliminate one-to-many correspondence. Second, we utilize the local geometry similarity between the input point cloud and permuted one to promote the learning of $\mathbf{P}$, i.e., neighbouring points in $\mathbf{A}$ (resp. $\mathbf{B}$) should also be neighbours in $\widehat{\mathbf{B}}$ (resp. $\widehat{\mathbf{A}}$), 
which is mathematically expressed as 
 \begin{align}
   \label{lmanifold}
 &\mathcal{L}_{mfd}(\mathbf{P})=\\
  &\sum_{i=1}^n\left(\sum_{k\in\Omega^a_i}\frac{\|\mathbf{p}_i\mathbf{B}-\mathbf{p}_k\mathbf{B}\|_2^2}{\|\mathbf{a}_i-\mathbf{a}_k\|_2^2}+\sum_{s\in\Omega^b_i}\frac{\|\mathbf{p}^i\mathbf{A}-\mathbf{p}^s\mathbf{A}\|_2^2}{\|\mathbf{b}_i-\mathbf{b}_s\|_2^2}\right), \nonumber
 \end{align}
where $\mathbf{p}_i$ (resp. $\mathbf{p}^i$)is the $i$-th row (resp. column) of $\mathbf{P}$, and $\Omega_i^a$ (resp. $\Omega_i^b$) is the index set of $k$ nearest neighbours of point $\mathbf{a}_i$ (resp. $\mathbf{b}^i$). 

Finally, the overall loss function for training CorrNet3D is written as 
\begin{align}
   \label{optimproblem}
\mathcal{L}&\left(\widetilde{\mathbf{A}}, \widetilde{\mathbf{B}},\mathbf{P}\right)= \\ &\mathcal{L}_{rec}\left(\widetilde{\mathbf{A}},\widetilde{\mathbf{B}} \right) +\lambda _1\mathcal{L}_{perm}\left( \mathbf{P} \right)  
+\lambda _2\mathcal{L}_{mfd}\left(\mathbf{P}\right), \nonumber
 \end{align}
where $\lambda_1$ and $\lambda_2>0$ are the parameters to balance the three terms. See Sec. \ref{ablation_} for the experimental validation towards such an unsupervised loss function. 

\subsection{Pseudo Clustering for Large-scale Point Clouds}\label{Pseudo}
As the size of predicted $\mathbf{P}$ depends on that of the input point cloud, directly inferring the correspondence of large-scale point clouds may cause a memory issue. To this end, we propose pseudo clustering, 
a simple yet effective 
approach. Specifically, during inference, we first apply a typical sampling method such as farthest point sampling (FPS) on input point clouds to sample a fewer number of points called key points, which are thought of as cluster centers and  fed into CorrNet3D, leading to the correspondence of the key points. 
Then the nearest neighboring points of each key point are found and sorted according to their Euclidean distances 
to center, and the correspondence of the neighbouring points of two corresponded key points are finally determined if two neighbouring points have the same rank in their own cluster. 
Such a pseudo clustering enables us to easily apply CorrNet3D to 
large-scale point clouds. 


It is also worth pointing out that such a simple strategy would degrade the performance of our method when directly applied on large-scale point clouds under the condition with sufficient memory to some extent; however, the experiment shows CorrNet3D can still predict more accurate correspondence than the method even trained with 3D meshes, demonstrating the strong ability of our CorrNet3D. 

\section{Experiments}
In this section, we conducted extensive experiments and comparisons 
on real scanned non-rigid shapes and synthetic non-rigid and rigid shapes to demonstrate superiority of CorrNet3D in both supervised\footnote{The supervised CorrNet3D (S-CorrNet3D) is achieved by removing the symmtric deformation module and training the remaining modules via minimizing the Euclidean distance between the predicted correspondence (i.e., $\mathbf{P}$) and the ground-truth one.} and unsupervised scenarios. 

\subsection{Experiment Setting}

\textbf{Datasets}. For non-rigid shape correspondence, we adopted Surreal~\cite{groueix20183d} as the training dataset, consisting of 230K samples, which were randomly grouped 
into 115K training pairs. We conducted the test on the SHREC dataset~\cite{donati2020deep}, which has 430 pairs of non-rigid shapes. For rigid shape correspondence, we adopted the training and test dataset splits of the Surreal dataset \cite{groueix20183d}, which  
contain 230K and 200 samples, respectively, and we randomly rotated and translated the samples 
to generate 230K pairs and 200 pairs for training and testing, respectively. Note that we chose these datasets in order to keep the same settings as the compared methods, including DeepGFM \cite{donati2020deep}, DCP \cite{wang2019deep} and RPMNet \cite{yew2020rpm}, for fair comparisons. For all the above training data, each point cloud contains 
1024 points.
The Surreal and SHREC datasets are both synthetic 3D meshes  and we randomly picked 1024 vertices to form the point clouds. 

\textbf{Metrics}. To fairly and quantitatively compare different methods, 
we define the corresponding percentage (Corr (\%)) to measure correspondence accuracy, i.e.,  \vspace{-0.25cm}
\begin{equation}
  \label{acc}
 \text{Corr}=\frac{1}{n}\left\|\widehat{\mathbf{P}}\odot \mathbf{P}_{gt}\right\|_1,
 \vspace{-0.25cm}
 \end{equation}
where $\odot$ is the Hadamard product of matrices, $\|\cdot\|_1$ is the $\ell_1$ norm of a matrix, and $\mathbf{P}_{gt}$ encodes the ground-truth correspondence.
Moreover, for a comprehensive comparison, we computed the corresponding percentage of different methods under various  tolerant errors defined as $r/dist_{max}$, where $dist_{max}:=\max\{\|\mathbf{a}_i-\mathbf{a}_j\|_2, \forall i, j\}$, 
and $r$ stands for the tolerant radius.
It is worth pointing out that the above quantitative evaluation criteria 
are similar to those used for 3D mesh-based shape correspondence~\cite{ovsjanikov2012functional,litany2017deep} 
which adopt the geodesic distance as the tolerant error 
requiring connectivity information. But such information is not available for point clouds, we directly compute the Euclidean distance. 

\textbf{Implementation details}. For the parameters in  Eq.~(\ref{optimproblem}), we empirically 
set $\lambda _1=0.1$ and $\lambda _2=0.01$. The shared symmetric deformer consists of a 3-layer MLP. We implemented it with 
the PyTorch framework \cite{paszke2017automatic} on GeForce RTX 2080Ti.We trained the models with Adam \cite{kingma2014adam} optimizer with the learning rate equal to 1e-4 and the batch size equal to 10 for 300 epochs.  


\subsection{Evaluation on Non-rigid Shapes}
\begin{figure}[t]
   \centering
   \includegraphics[width=0.4\textwidth]{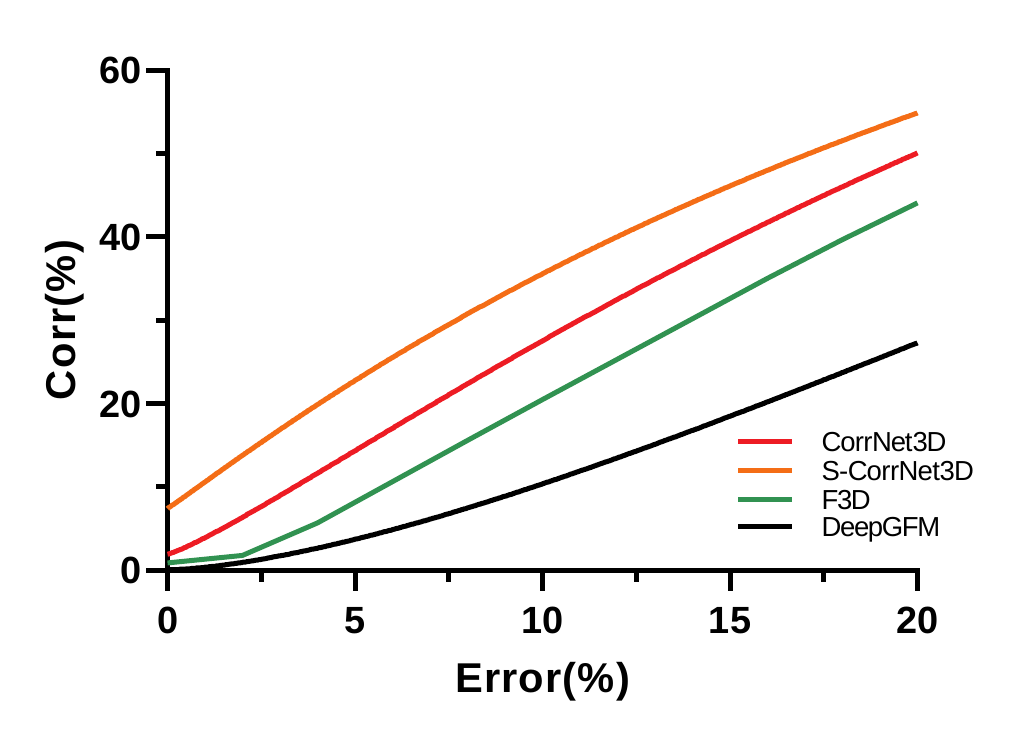}
   \caption{Quantitative comparisons of different methods for non-rigid shape correspondence.}
     \label{Nonrigid}
 \end{figure}
 \begin{figure}[t]
  \centering
  \includegraphics[width=0.4\textwidth]{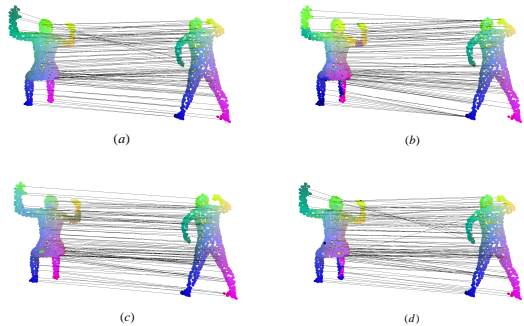}
  \caption{ Visual comparisons of different methods for non-rigid shape correspondence in the form of point clouds. 
  Both lines and colors are used for illustrating the correspondence.  (a) Ground-truth. (b) F3D. (c) CorrNet3D. (d) S-CorrNet3D.}
  \label{ourNonrigidResults}
 \end{figure} 

In this scenario, we compared CorrNet3D and supervised CorrNet3D (S-CorrNet3D) 
with unsupervised F3D\footnote{We obtained the unsupervised F3D by modifying the supervised FlowNet3D~\cite{liu2019flownet3d}, i.e., we only replaced the loss function with the Chamfer Distance. 
} ~\cite{liu2019flownet3d} and supervised DeepGFM~\cite{donati2020deep} taking 3D meshes as input.

\begin{figure}[t]
   \centering
   \includegraphics[width=0.4\textwidth]{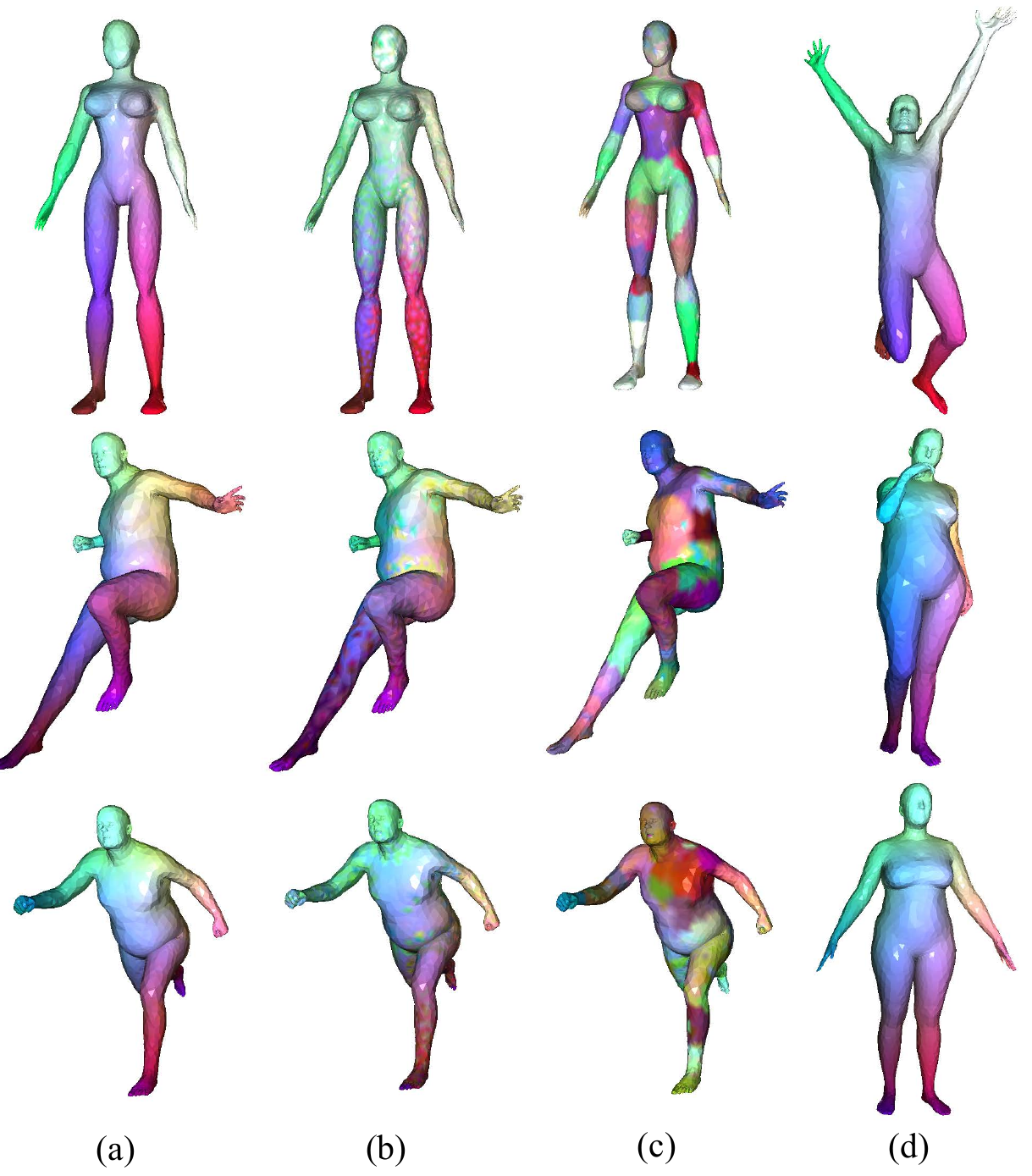}
   \caption{Visual comparisons of different methods on non-rigid shapes in the form of 3D meshes. 
   Each mesh contains 5200 vertices. We render the two corresponded points of the two shapes in the same color.  (a) and (d) show the ground-truth correspondence. (b) and (c) show the results of CorrNet3D and DeepGFM, respectively. Note that CorrNet3D takes only 1024 vertices as input and DeepGFM takes 3D meshes as input. We compute the correspondence of the remaining points is obtained via a pseudo clustering strategy.  
   }
   \label{Dense}
 \end{figure}

Fig.~\ref{Nonrigid} shows the quantitative comparisons of different methods, 
where it can be observed that S-CorrNet3D always produces the best performance, and both CorrNet3D and S-CorrNet3D  
consistently outperform 
F3D and DeepGFM. Especially, 
the performance advantage of our methods over DeepGFM is more obvious 
with the tolerant error increasing. 
Fig.~\ref{ourNonrigidResults} shows visual comparisons of F3D, CorrNet3D and S-CorrNet3D on point clouds, 
where the predicted correspondence is visualized with colors and lines. 
From Fig.~\ref{ourNonrigidResults}, we can see that CorrNet3D and S-CorrNet3D produce more accurate correspondence than F3D, especially  at feet, hands and the right leg. 
Fig.~\ref{Dense} shows the visual comparisons of CorrNet3D and DeepGFM on 3D meshes,  
which further demonstrates our method's advantage.
That is, the predicted correspondence by our CorrNet3D is closer to the ground-truth one. However, 
DeepGFM results in patchy distributed wrong correspondence,
although it utilizes additional connectivity information and 
ground-truth correspondence as supervision.


\begin{figure}[t]
  \centering
  \includegraphics[width=0.45\textwidth]{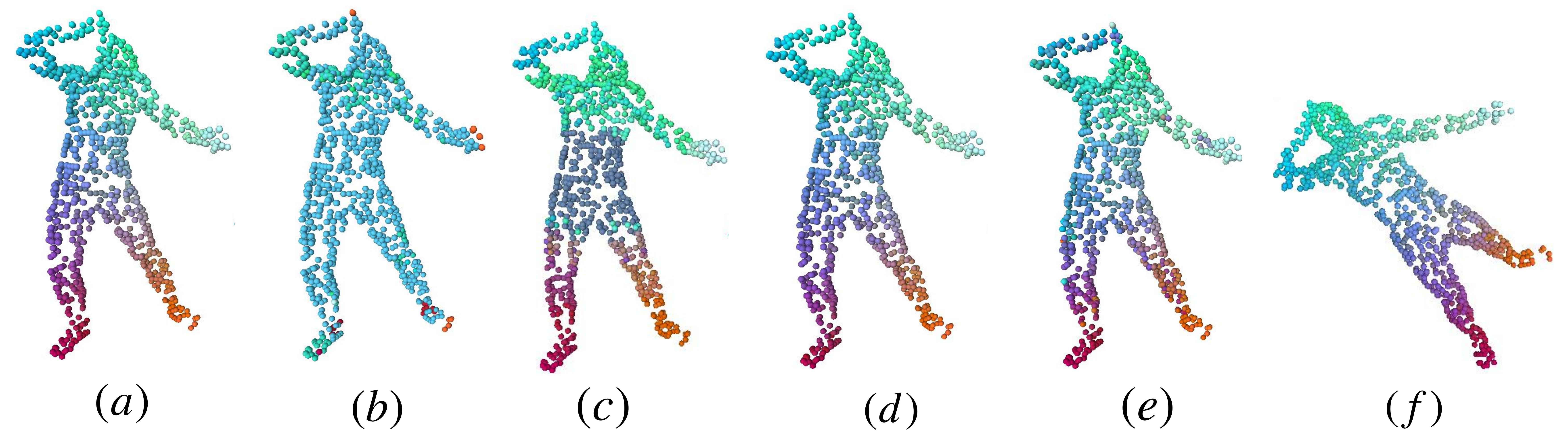}
 \caption{Visual comparisons of different methods for rigid shape correspondence. 
 (a) and (f) show the ground-truth correspondence. (b),(c),(d) and (e) show the results of DCP, RPMNet, CorrNet3D and S-CorrNet3D, respectively. 
  }
  \label{OurRIGIDRESULTS}
 \end{figure}
 
\begin{figure}[t]
   \centering
   \includegraphics[width=0.4\textwidth]{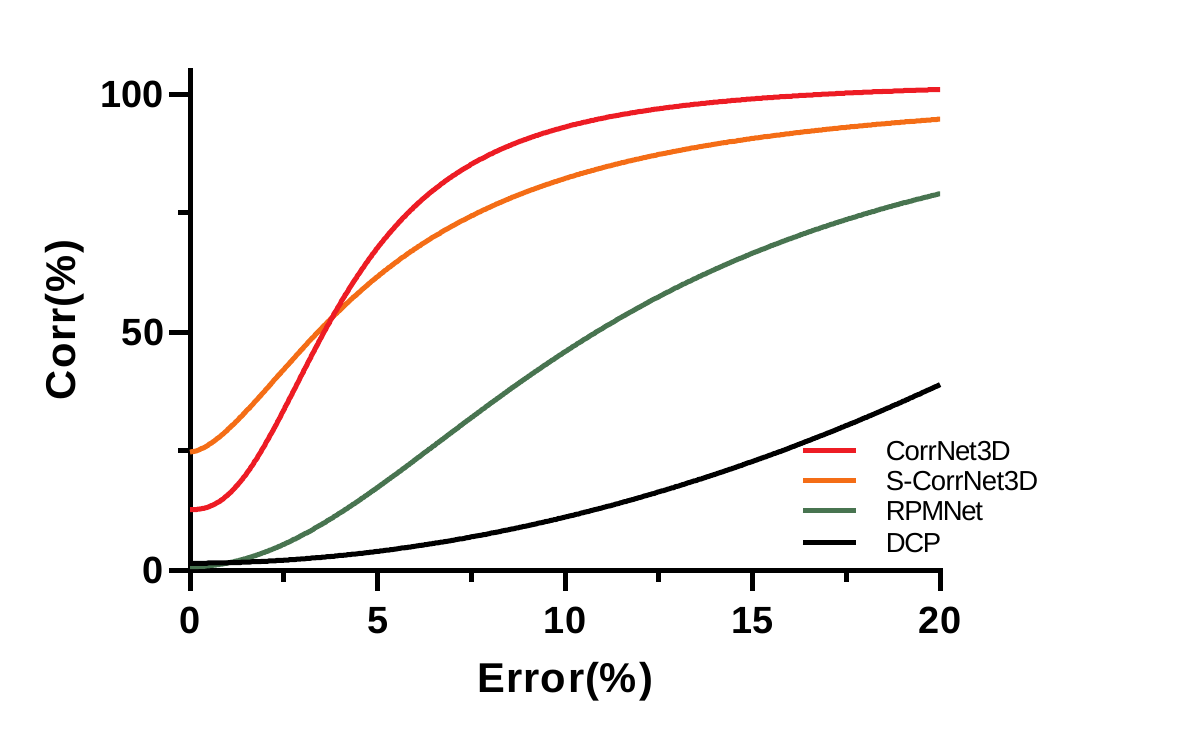}
   \caption{Quantitative comparisons of different methods for rigid shape correspondence.}
     \label{Rigid}
 \end{figure}
\subsection{Evaluation on Rigid Shapes}
In this scenario, we compared our CorrNet3D and S-CorrNet3D with DCP \cite{wang2019deep} and RPMNet \cite{yew2020rpm}. Notice both DCP and RPMNet require ground-truth rigid transformation as supervision during training\footnote{As DCP and RPMNet are able to generate soft correspondence by transformer and Sinkhorn, respectively, 
We output the soft correspondence matrix and set each row's max value as 1, and others as 0 to get a binary correspondence matrix for comparison.}.
Fig.~\ref{Rigid} reports the quantitative comparisons of different methods, 
where it can be seen that  CorrNet3D and S-CorrNet3D consistently outperform DCP an RPMNet. 
Interestingly but not surprisingly,  the unsupervised CorrNet3D 
performs even better than the other three supervised methods.  
The reason is that 
the freedom and searching space for the model in the supervised manner will be limited by the training dataset, making it harder to adapt the trained model to data with large transformation. 
Fig.~\ref{OurRIGIDRESULTS} visually compares the results of different methods, 
where it can be observed that DCP even fails to obtain correct matching, and RPMNet cannot predict correct matching for hands and the body part. In contrast, S-CorrNet3D and CorrNet3D are able to generate more accurate matching results, which are closer to ground-truth ones.



\subsection{Evaluation on Real Scanned Data}
\begin{figure}[t]
   \centering
   \includegraphics[width=0.4\textwidth]{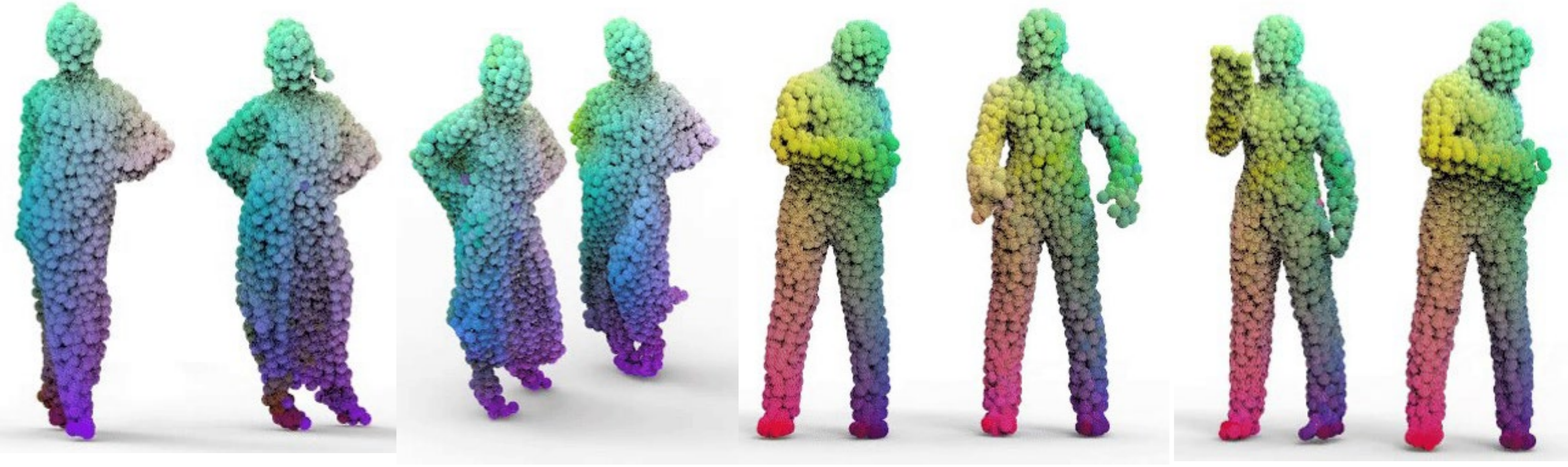}
   \caption{Visual results of CorrNet3D on 8iVFB \cite{d20178i} - a real scanned dataset.
   }
     \label{Real}
 \end{figure}

We also examined the robustness of CorrNet3D on a real scanned dataset, i.e., 8iVFB \cite{d20178i}, including the dynamic point cloud sequences of human motion with serious deformation. The test point clouds contain 1024 points each randomly picked from the original ones. 
As illustrated in Fig.~\ref{Real}, where the two corresponding points of two shapes predicted by CorrNet3D are visualized with the same color,  
we can observe CorrNet3D trained on the synthetic dataset 
still produces impressive performance on real data even with  serious deformation, demonstrating the CorrNet3D's strong ability. 

\subsection{Ablation Study}\label{ablation_}
In this section, we conducted extensive ablation studies for a comprehensive understanding towards our CorrNet3D. 
We carried out experiments on non-rigid Surreal dataset in the unsupervised scenario. 

\begin{figure}[t]
  \centering
  \includegraphics[width=0.40\textwidth]{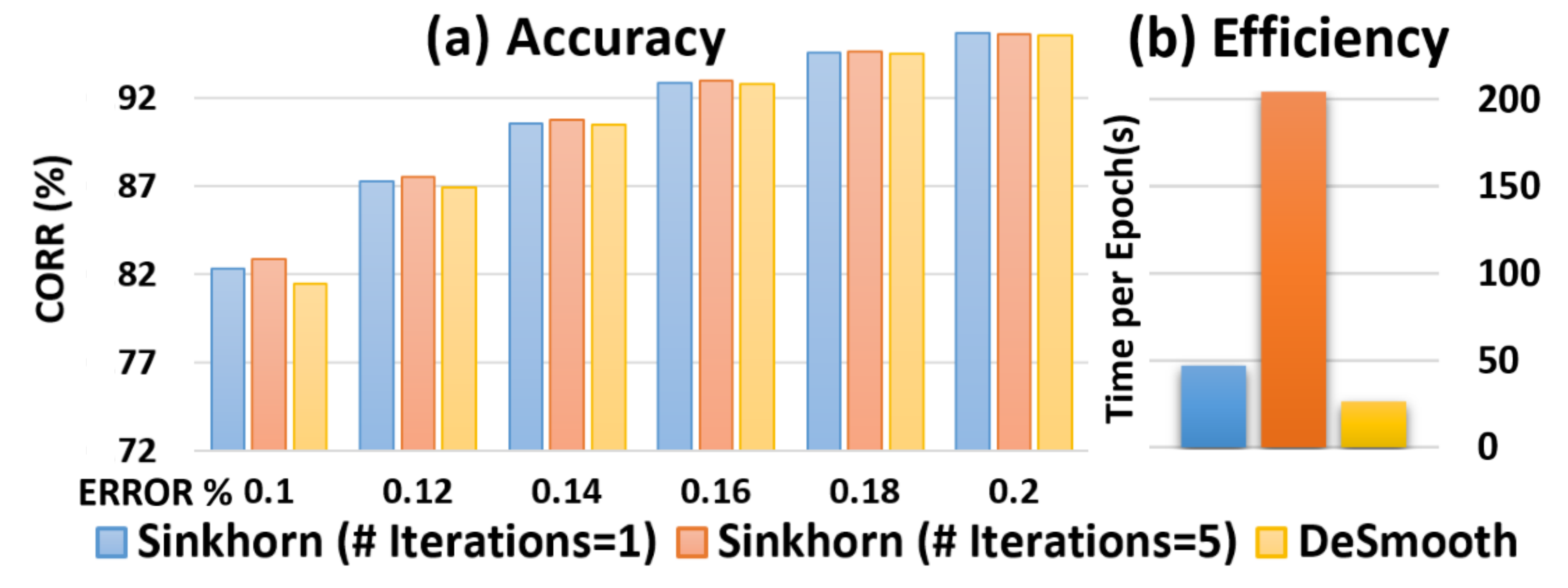}
  \caption{Comparisons between our DeSmooth and the Sinkhorn layer in terms of accuracy (a) and efficiency (b).}
  \label{sinkhorn2}
 \end{figure}

\textbf{DeSmooth module}. We compared our DeSmooth with the Sinkhorn layer. Specifically, we replaced the DeSmooth module of CorrNet3D with the Sinkhorn layer, while keep all the remaining settings the same. 
As shown in Fig.
~\ref{sinkhorn2}, it can be observed that 
under the same tolerant error, 
the accuracy of our DeSmooth is on par with that of the Sinkhorn layer, 
while our DeSmooth improves the efficiency up to 8$\times$.



\textbf{Deformation module.} The results listed in  Table~\ref{table:decoder} validate the effectiveness of our symmetric deformer design and its deformation-like behavior. 
That is, The correspondence accuracy of the other two settings, i.e., (a) replacing the symmetric deformer of CorrNet3D with fully connected (FC) layers and (b) training CorrNet3D without (w/o) feeding global features into the symmetric deformer (i.e., the absence of the shape information),  
decreases significantly.
We also compared the symmetric deformer w/ and w/o shared parameters 
As listed in Table \ref{table:decoder}, comparable performance is achieved under these two settings, but the deformer w/ shared parameters is more memory-efficient as the number of parameters reduces by half.

\begin{table}[t]
  \centering
  \caption{Validation of the effectiveness and behavior of the symmetric deformer.
  Here the tolerance error is equal to 20$\%$.}
  \label{table:decoder}
  \footnotesize
  \begin{tabular}{c||c|c|c|c}
    \toprule[1.5pt]
  Module & (a) Fully & (b) Deformer & (c) Deformer  & (d) Deformer   \\ 
        &  connected  & (\textit{w/o} $\mathbf{v}_a,\mathbf{v}_b$) &  (not shared)  & (shared) \\ \hline
  Corr($\%$)    & 26.21        &  25.24    & 95.97                    & 95.61               \\\bottomrule[1.5pt]
  \end{tabular}
  \end{table}

\textbf{Loss function}. In Table~\ref{table:loss}, we compared the performance of CorrNet3D when trained with different loss functions. 
The effectiveness of the regularization terms can be validated by comparing the 2$^{nd}$ and 4$^{th}$ columns, and the advantage of our Euclidean distance-based reconstruction loss over the Chamfer Distance can be demonstrated by comparing the 3$^{rd}$ and 4$^{th}$ columns. 


\begin{table}[t]
  \centering
  \caption{Validation of the effectiveness of loss setting in Eq. (\ref{optimproblem}),   $\mathcal{L}_{reg}=\lambda_1\mathcal{L}_{perm}+\lambda_2\mathcal{L}_{mfd}$. 
  The tolerance error is 20$\%$.}
  \label{table:loss}
  \begin{tabular}{c||c|c|c}
\toprule[1.5pt]
  Loss       &    $\mathcal{L}_{rec}$ &   $\mathcal{CD}+\mathcal{L}_{reg}$&   $\mathcal{L}_{rec}+\mathcal{L}_{reg}$ \\ \hline
  Corr($\%$) & $\quad 31.49 \quad$    & 25.74       &95.61                  \\ 
\bottomrule[1.5pt]
  \end{tabular}
\end{table}


 \begin{figure}[t]
  \centering
  \includegraphics[width=0.4\textwidth]{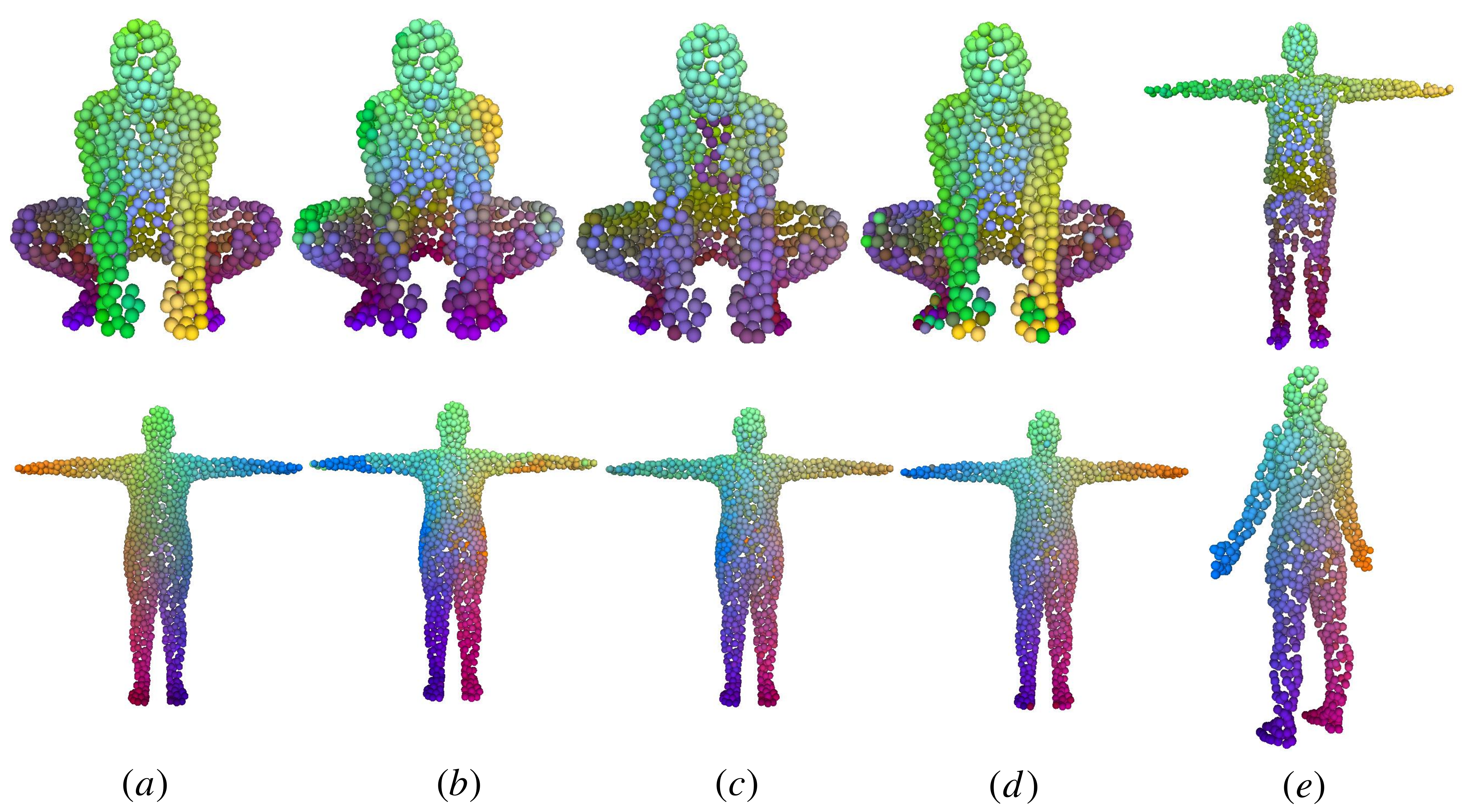}
  \caption{Visualization of two 
  failure cases of non-rigid shape correspondence. 
  (a) and (e) show the ground-truth correspondence. 
  (b) and (e) show the results by F3D. (c) and (e) show the results by CorrNet3D. (d) and (e) show the results by S-CorrNet3D. 
  The corresponding points are shown as the same color. Note that the two unsupervised methods F3D and CorrNet3D fail, while the supervised S-CorrNet3D can obtain correct results.}
  \label{nonrigid_failure_cases}
  \vspace{-0.4cm}
 \end{figure}

\textbf{Failure cases}. 
Here we present two failure cases, which occur for computing non-rigid shape correspondence on highly symmetric and distorted shapes.  
As show in Fig.~\ref{nonrigid_failure_cases},  we can see that the two unsupervised methods i.e., F3D and CorrNet3D, both generate wrong correspondence 
especially in the hands and feet. However, 
the supervised S-CorrNet3D can successfully obtain the correct correspondence. In future,  solving such issues without using heavy annotations or simple data augmentation is a promising direction. 
\section{Conclusions}
We have presented the first unsupervised and end-to-end learning framework named CorrNet3D for building dense correspondence between 3D shapes in the form of point clouds. Unlike existing works, we addressed this challenging problem from the deformation-like reconstruction perspective.  Note that CorrNet3D is a flexible framework in that it can be simplified to work in a supervised manner when annotated data are available.
We demonstrated the significant advantages of our methods over state-of-the-art ones by conducting extensive experiments on real scanned and synthetic data including rigid and non-rigid shapes in both unsupervised (CorrNet3D) and supervised (S-CorrNet3d) scenarios, as well as comprehensive ablation studies. 
We believe our methods will bring benefits to other tasks, such as point cloud sequence compression which needs correspondence for eliminating the inter-frame redundancy, and deep learning-based point cloud sequence analysis, which usually has to align points from different frames for feature aggregation.  



{\small
\balance
\bibliographystyle{ieee_fullname}
\bibliography{cvpr}
}

\end{document}